%% file: final-hmm-main-v0.tex
\documentclass[10pt, onecolumn]{IEEEtran}

\input{test.tex}

\usepackage{enumitem}
\usepackage[margin=1in]{geometry}
\usepackage{cite}
\usepackage{algorithmic}
\usepackage{framed}

\newcommand{\rvd}[1]{{\color{black}{#1}}}
\newcommand{\rve}[1]{{\color{black}{#1}}}
\newcommand{\rvf}[1]{{\color{black}{#1}}}

\newcommand{\rva}[1]{{\color{black}{#1}}}
\newcommand{\rvb}[1]{{\color{black}{#1}}}
\newcommand{\rvc}[1]{{\color{black}{#1}}}

\newcommand{\qq}[1]{\color{Red}{#1}}

\begin{document}

\title{ Minimal Realization Problems for Hidden Markov Models}
\author{
  Qingqing Huang$^{\dagger}$, Rong Ge$^{\ddagger}$, Sham Kakade$^{\ddagger}$, Munther
  Dahleh$^{\dagger}$
  \thanks{Copyright (c) 2015 IEEE. Personal use of this material is permitted. However, permission
    to use this material for any other purposes must be obtained from the IEEE by sending a request
    to pubs-permissions@ieee.org.}
  \thanks{Q. Huang, and M. Dahleh are with the Laboratory of Information and
  Decision Systems, Massachusetts Institute of Technology (Emails: \{qqh,
  dahleh\}@mit.edu); R.Ge and S.Kakade are with Microsoft Research, New
  England (Emails: \{rongge, skakade\}@microsoft.com).  }
}
\date{}
\maketitle

\begin{abstract}

  \rva{

    This paper addresses two fundamental problems in the context of \rvd{Hidden Markov
      Models} (\rve{HMMs}). The first problem is concerned with the characterization and
    computation of a minimal order HMM that realizes the exact joint densities of an
    output process based on only finite strings of such densities (known as HMM partial
    realization problem).
    %
    The second problem is concerned with learning a HMM from finite output observations of
    a stochastic process.
    We review and connect two fields of studies: realization theory of HMMs, and the recent
    development in spectral methods for learning latent variable models.
    Our main results in this paper focus on generic situations, namely, statements that will be true
    for almost all HMMs, excluding a measure zero set in the parameter space.
    %
    %
    In the main theorem, we show that both the minimal quasi-HMM realization and the minimal HMM
    realization can be efficiently computed based on the joint probabilities of length $N$ strings,
    for $N$ in the order of $\mc O(\log_d(k))$. In other words, learning a quasi-HMM and an HMM have
    comparable complexity for almost all HMMs.
}

\end{abstract}


\section{Introduction}
\subsection{Background}
Hidden Markov Models (HMMs) are widely used for describing discrete random processes, especially in
the applications involving temporal pattern recognition such as speech and gesture recognition,
part-of-speech tagging and parsing, and bioinformatics.
The Markovian property of the hidden state evolution potentially \rve{leads} to a low complexity
representation of the output random process.
In this work, we consider the long-standing HMM realization problem: given some partial knowledge
about the output process of an unknown HMM, can we generalize it to a full description of the random
process?

Consider a discrete random process $\{y_t:t\in\mbb Z\}$, which assumes values in a finite alphabet
$[d]\equiv\{1,\cdots,d\}$. Assume that $y_t$ is the output process of a stationary HMM of finite
order.
Let the random vector $\mb y_1^N = (y_1,\dots,y_N)$ denote an string of length $N$, which
assumes values in the $N$-ary Cartesian product $[d]^N$.
The process $y_t$ is fully characterized by the joint probabilities of strings of any length in
\rvc{the countably infinite} table (denoted by $\mc P^{(\infty)}$):
\begin{align*}
  \Big\{\mbb P(y_1 = l_1, \cdots, y_N = l_N): \fa \mb l_1^N\in[d]^N, \fa
  N\in\mbb Z\Big\}.
\end{align*}
There are three main concerns in the realization problem:
\begin{enumerate}[leftmargin=*]
\item \tb{(Informational complexity)} Suppose that the underlying HMM is of order $k$, and we are
  given the joint probabilities of all the length $N$ strings, namely:
  \begin{align*}
    \mc P^{(N)} \equiv\Big\{\mbb P(y_1 = l_1, \cdots, y_N = l_N): \fa \mb l_1^N\in[d]^N\Big\},
  \end{align*}
 how large \rvc{does $N$ need} to be so that we can compute
  $\mc P^{(\infty)}$ based on  $\mc P^{(N)}$?

\item \tb{(Computational complexity)}
  Can we solve the realization problem with runtime polynomial in the dimensions (alphabet size $d$
  and order of the underlying HMM $k$)?

\item \tb{(Statistical complexity)} When $\mc P^{(N)}$ is estimated from sample sequences and has
  some  estimation error, are the realization algorithms robust to the input errors?

\end{enumerate}

These are long standing  questions, and there are several lines of work within different
communities at tempting to address these questions.
It has long been known that, in the information theoretic sense, there exist hard cases of HMMs that
are not efficiently PAC learnable \cite{kearns1994learnability} \cite{mossel2005learning}. However,
a more practical question is\rvc{,} can we efficiently solve the realization / learning problem
\rvc{for most HMMs}?
In this work, we focus on generic analysis and show that, for almost all HMMs, i.e., excluding those
whose parameters are in a measure zero set
\footnote{
  In our setting, algebraic genericity coincides with the measure theoretic notion of
  generic. Throughout the discussion, for fixed alphabet size $d$ and order $k$, we call an HMM in
  general position if its transition and observation matrix are in general position, which is
  equivalent to ``almost everywhere in the parameter space of $\{Q\in\R^{k\times
    k}_+,O\in\R^{d\times k}_+: \mb e^\top Q = \mb e^\top, \mb e^\top O = \mb e^\top\}$''.  },
\rvf{the} realization problems can be efficiently solved with poly time algorithms.

\subsection{Organization}
\rvd{To study the HMM realization problems, we focus on algorithms based on \rvc{the spectral
  method}.}
The basic idea  is to exploit the recursive structural properties of the underlying finite state
model, and write the joint probabilities in $\mc P^{(N)}$ into a specific form which admits
\emph{rank decomposition}, where the rank reveals the minimal order of the realization and the model
parameters can be extracted from the factors.
%

In the first part (Section~\ref{sec:minimal-quasi-hmm}), we consider the problem of finding the
minimal quasi-HMM realization.
Quasi-HMMs are associated with different names in different communities, for example finite state
regular automata \cite{bailly2011quadratic,Balle2013spectral}, regular quasi realization
\cite{vidyasagar2011complete,mossel2005learning}, and operator models \cite{mossel2005learning,
  hsu2012spectral}.  We mostly follow the terminologies in \cite{vidyasagar2011complete}.
\rva{Algorithm~\ref{alg:quasi} is the well-known algorithm for finding the minimal order quasi-HMM
  realization (to be rigorously defined later). However, in general the window size $N$ can not be
  specified a priori and thus the complexity of the algorithm cannot be explicitly \rvd{determined}.
  In Theorem~\ref{thm:window-size}, we show that, if the output process is generated by an general
  position HMM with \rvf{order $k$}, we only need the window size $N$ in the order of $ \mc
  O(\log_d(k))$ for pinning down $\mc P^{(\infty)}$ based on $\mc P^{(N)}$, where $d$ is the output
  alphabet size.
  Moreover, we show that Algorithm~\ref{alg:quasi} has runtime and sample complexity both polynomial
  in the relevant parameters.}%

In the second part (Section~\ref{sec:minim-hmm-real}), we consider the problem of finding
the minimal HMM realization, using tensor decomposition methods, which rely on the
uniqueness of tensor decomposition to \rvf{recover the minimal order HMM that is unique up
  to hidden states permutation}.
Tensor decomposition based algorithms for learning HMMs are studied in \cite{anandkumar2012tensor,
  allman2009identifiability,bhaskara2013uniqueness}.
In these works, the transition matrix is always assumed to be of full rank.
Similar to that in the quasi-HMM realization problem, in general the required window size $N$ and
also the complexity of the algorithm cannot be determined a priori.
In \cite{allman2009identifiability}, the authors examined the generic identifiability conditions of
HMM, and showed that generically it suffices to pick the window size $N=2n+1$ \rvc{for some positive
  integer $n$,} such that ${ n+d-1 \choose d-1 }\ge k$. In the case where $d$ is much smaller than
$k$, $n$ needs to be in the order of $\mc O(k^{1/d})$.
Another bound on the window size $N$ is given in \cite{bhaskara2013uniqueness}, which is in the order
of $\mc O(k/d)$.
However, the size of the tensor in the decomposition is exponential in $n$, thus all
these bound lead to runtime exponential in $k$.


\rva{In Section~\ref{sec:minim-hmm-real}, we propose a two-step realization approach, and analyze the
identifiability issue of the two steps. Then, we show that for the processes generated by almost all
HMMs, the window size $N$ only needs to be in the order of $\mc O(\log_d(k))$ for finding the
minimal HMM realization. This means that for most HMMs, finding minimal quasi-HMM and minimal HMM
realizations are actually of equal difficulty.
}

\section{ Minimal realization problem formulation}
\label{sec:2}

In this section. we first review the basics of HMMs, and then formally introduce the
quasi-HMM and HMM realization problems.

\subsection{Preliminaries on HMMs}
An HMM determines the joint probability distribution over sequences of hidden states $\{x_t: t\in
\mbb Z\}$ and observations $\{y_t: t\in\mbb Z\}$.
For simplicity, we call each output $y_t$ as a ``letter'' taking value from some discrete alphabet
$[d]$, and a sequence of $n$ letters is referred to as a ``string'', taking value from the Cartesian
product $[d]^n$. \rvc{We use $[d^N]\equiv\{1,\dots, d^N\}$ to denote the vectorized indices in
  $[d]^n$.}

The joint distribution of $\{x_t,y_t:t\in\mbb Z\}$ from a stationary HMM is parameterized by a
pair \rve{of} matrices: the state transition matrix $Q\in\R_+^{k\times k}$, and the
observation matrix $O\in\R_+^{d\times k}$, which satisfy $ \mb e^\top O = \mb e^\top$ and
$ \mb e^\top Q = \mb e^\top$, where $\mb e$ is the all ones vector.
The hidden state $x_t$ evolves following a Markov process:
\begin{align*}
  \mbb P(x_{t+1} = j| x_{t} = i) = Q_{j,i}.
\end{align*}
Let $\pi$ denote the stationary state distribution, i.e., $\pi_i = \mbb P[x_t = i]$ and $Q\pi =
\pi$.  Without loss of generality, we assume that $\pi_i>0$ for all $i\in[k]$.
We also define the backward transition matrix $\wt Q\in\mbb R^{k\times k}$:
\begin{align*}
  \mbb P(x_{t-1}=j|x_{t} = i) = \wt Q_{j,i}.
\end{align*}
Observe that the matrix $\wt Q$ is related to $Q$ as: $ \wt Q = Diag(\pi) Q^\top Diag(\pi)^{-1}$.
Conditioned on the hidden state taking value $i$, the probability of observing letter $j$ is:
\begin{align*}
  \mbb P(y_t = j| x_t= i) = O_{j,i}.
\end{align*}
\rvf{We call two HMMs equivalent if the output processes are statistically indistinguishable.}

\rva{The order of the HMM is defined to be the number of hidden states, denoted by $k$.}
We will denote the class of all HMMs with output alphabet size $d$ and order $k$ by $
\Ta_{(d,k)}^h$.

\subsection{Problem formulations}
\label{sec:problem-formulations}

\rva{The \emph{ realization problem} takes as inputs the probabilities of finite length strings
  for a fixed window size $N$ ($\mc P^{(N)}$), and finds a finite state model of the minimal order
  to describe the entire output process ($\mc P^{(\infty)}$).
We aim to find the most succinct description of the process, namely the minimal order realization,
where the ``order'' refers to the number of states of the underlying finite state model.
Without loss of generality, we assume that the process has a minimal realization of
\rvf{\emph{order $k$}} and examine under what conditions the algorithms can recover an equivalent
minimal order realization.  }

Next, we introduce two classes of finite state models, both of which can realize an HMM output process.

\begin{definition}[Quasi-HMM realization \cite{vidyasagar2011complete}]
  \label{def:quasi-hmm}
  Let $\ta^o$ be a tuple: $ \ta^o = ( k, u,v \in\mbb R^{k}, A^{(j)}\in\mbb R^{k\times k}: \fa
  j\in[d]).$ We call $\ta^o$ a quasi-HMM realization of order $k$ for a stationary process
  $\{y_t:t\in\mbb Z\}$ if the three conditions hold: ($\fa \mb l_1^N\in[d]^N, \fa N\in\mbb Z$)
  \begin{align}
    &\mbb P(\mb y_1^N = \mb l_1^N) = u^\top A^{(l_1)}A^{(l_2)}\cdots A^{(l_N)} v,
    \label{eq:recur-quasi}
    \\
    & u^\top(\sum_{j=1}^{d} A^{(j)}) = u^\top,
    \label{eq:norm-quasi-1}
    \\
    & (\sum_{j=1}^{d} A^{(j)} ) v= v.
    \label{eq:norm-quasi-2}
  \end{align}
\end{definition}

\begin{definition}[Equivalent quasi-HMM realizations]
  Two quasi-HMM realizations $\ta^o= (k,u, v, A^{(j)}:j\in[d])$ and $\wt\ta^o = (k, \wt u, \wt
  v, \wt A^{(j)}:j\in[d])$ are called equivalent, if there is a full rank matrix $T\in \mbb
  R^{k\times k}$ such that:
  \begin{align*}
    \wt u = \rvc{T^\top u},\ \wt v = T^{-1}v,\ \wt A^{(j)} = T^{-1} \rva{ A^{(j)}} T, \quad \fa j\in[d].
  \end{align*}
\end{definition}


\begin{definition}[HMM realization]
  Let $\ta^h$ be a tuple: $\ta^h = ( k, O\in\mbb R_+^{d\times k}, Q\in\mbb R_+^{k\times k})$.  We
  call $\ta^h$ an HMM realization of order $k$ for a stationary random process $\{y_t:t\in\mbb Z\}$,
  if the matrices $Q$ and $O$ are column stochastic, and the output process of the HMM defined by
  the transition matrix $Q$ and observation matrix $O$ has the same distribution as $y_t$.
\end{definition}


HMM realizations are in a subset of the model class of quasi-HMM realizations. Given an HMM
realization $\ta^{h}=( k, O, Q)$, one can construct the following quasi-HMM realization
$\ta^o=(k,u,v,A^{(j)}:j\in[d])$:
  \begin{align}
    &u = \mb e,
    \label{eq:o-to-h-u}
    \\
    &v = \pi,
    \label{eq:o-to-h-v}
    \\
    &A^{(j)} = Q Diag(O_{[j,:]}),\quad  \fa j\in[d].
    \label{eq:o-to-h-A}
  \end{align}

  The minimal \rvc{(quasi-)}HMM realization problem is formally stated below: Assume that the random
  process is the output of an HMM of \rvf{order $k$}. \rvc{How large does the window size
    $N$ need to} be, so that given the joint probabilities $\mc P^{(N)}$ we can efficiently
  construct a minimal \rvc{(quasi-)}HMM realization for the process?

\section{Minimal Quasi-HMM Realization}
\label{sec:minimal-quasi-hmm}
In this section, we address the minimal quasi-HMM realization problem.
We first review the widely used algorithm\cite{anderson1999realization,Balle2013spectral}; then we
show for HMMs in general position, the window size $N$ only needs to be in the order of $\mc
O(\log_d(k))$ to guarantee the correctness of the algorithm; we also give an example of hard case
(degenerate) which needs $N$ to be as large as $k$; finally we examine the stability of the
algorithm.

\subsection{Algorithm }

\rvc{ For notational convenience, we define the bijective mapping $L: [d]^n\to [d^n]$ which maps the
  multi-index $\mb l_1^N = (l_{1},\cdots,l_n)\in [d]^n$ to the index $ L(\mb l_1^n) = (l_1-1)d^{n-1}
  + (l_2-1)d^{n-2} + \cdots + l_n \in[d^n].$}

Given the length $N$ joint probabilities $\mc P^{(N)}$, where $N=2n+1$ for some positive number $n$,
we form two matrices $H^{(0)}, H^{(j)} \in \mathbb R^{d^n\times d^n}$ for all $j\in[d]$ as below:
\begin{align}
  \label{eq:H0-def}
  [ H^{(0)}]_{L(\mb l_1^n),L(\mb l_{-1}^{-n})}& = \mbb P\Big(\mb y_{-1}^{-n} = \mb l_{-1}^{-n},\ \mb y_0^{n-1} =
  \mb l_1^n\Big),
  \\
  \label{eq:Hj-def}
  [H^{(j)}]_{L(\mb l_1^n),L(\mb l_{-1}^{-n})} &= \mbb P\Big(\mb y_{-1}^{-n} = \mb l_{-1}^{-n}, y_0 = j, \mb y_1^{n} =
  \mb l_1^n\Big),
\end{align}
where $\mb l_1^n=(l_1,\dots, l_n)$ and $\rvc{\mb l_{-1}^{-n}}=(l_{-1}, l_{-2},\dots, l_{-n})
\in[d]^n$ denotes the length $n$ string corresponding to the future and the past $n$ time slots,
respectively.
Note that the ``future'' observations and the ``past'' observations are independent conditioned on
the ``current'' state, \rve{which is the Markovian property that Algorithm~\ref{alg:quasi} relies on.}

\begin{algorithm}[H]
  \caption{Minimal  quasi-HMM realization
  }
  \label{alg:quasi}
  \begin{algorithmic}
    \REQUIRE {$H^{(0)}, H^{(j)}\in\mbb R^{d^n\times d^n}$ for all $j\in[d]$}

    \ENSURE { $\wt \ta^{o} = ( k, \wt u, \wt v,  \wt A^{(j)}: j\in[d] )$}
    \begin{enumerate}[leftmargin=*]

      \STATE Compute the SVD of $H^{(0)}$:
      \begin{align}
        \label{eq:svd-H0}
           &H^{(0)} = U_HD_HV_H'.
         \end{align}
         Set $U = U_HD_H^{1/2},\quad V = V_HD_H^{1/2}$.

         \STATE Let $\wt k$ be the rank of $H^{(0)}$, and let
         \begin{align}
           \label{eq:find-wt-uv}
           \wt u = U' \mb e, \quad \wt v = V' \mb e.
         \end{align}

         \STATE Let $U^\dag$ and $V^\dag$ be the pseudo inverse of $U$ and $V$.
         \begin{align}
           \label{eq:find-wt-Aj}
           \wt A^{(j)} = U^{\dagger}H^{(j)}(V^{\dagger})', \quad \fa j \in [d].
         \end{align}

    \end{enumerate}
  \end{algorithmic}
\end{algorithm}

\rvb{The core idea of Algorithm~\ref{alg:quasi} was discussed in \cite{ito1992identifiability}, and
  it has been rediscovered numerous times in the literature in slightly different forms
  \cite{anderson1999realization,Balle2013spectral}. }We summarize the main idea below.

\begin{remark}[Minimal order]
Let $\ta^o=(k,u,v,A^{(j)}:j\in[d ])$ be a {\em minimal} quasi-HMM realization \rvf{of order $k$} for the
process considered.  Since the joint probabilities can be factorized in terms of the $A^{(j)}$'s as
in \eqref{eq:recur-quasi}, one can factorize $H^{(0)}$ and $H^{(j)}$'s as below:
\begin{align*}
  H^{(0)} = EF^\top,\quad    H^{(j)} = EA^{(j)}F^\top,
\end{align*}
where the matrices $E,F\in \R^{d^n\times k}$ are functions of $\ta^o$. In particular, the $L(\mb
l_1^n)$-th row of $E$ and $F$ are given by:
\begin{align}
  \label{eq:E-def}
  &E_{[L(\mb l_1^n),:]} = u^\top (A^{(l_n)}\cdots A^{(l_1)}),
  \\
  \label{eq:F-def}
  &F_{[L(\mb l_1^n),:]} = v^\top (A^{(l_n)}\cdots A^{(l_1)})^\top.
\end{align}
Note that if both $E$ and $F$ have full column rank $k$, then $H^{(0)}$ has rank $k$, according to
Sylvester's inequality.  \rva{Any rank factorization leads to an equivalent minimal quasi-HMM
  realization \rvf{of order $k$}. The minimal order condition, though not explicitly enforced, is
  reflected in the \emph{rank} factorization, as any quasi-HMM realization of lower order results in
  a matrix $H^{(0)}$ of lower rank, which leads to a contradiction.}

\end{remark}

The correctness of the algorithm crucially relies on matrix $H^{(0)}$ achieving its maximal rank
$k$, which equals the order of the minimal realization. A necessary condition for the correctness of
the algorithm is stated below.

\begin{lemma}[Correctness of Algorithm \ref{alg:quasi}]
  \label{prop:correct-quasi-alg}
  \rvb{ Assume the process \rvf{has a minimal  quasi-realization $\ta^o$ of order $k$}.  Algorithm
    \ref{alg:quasi} returns a \emph{minimal} quasi-HMM realization $\wt\ta^o$ that is equivalent to
    $\ta^o$, if the matrices $E$, $F$ defined in \eqref{eq:E-def} and \eqref{eq:F-def} have full
    column rank $k$.}
\end{lemma}

Increasing the window size $N$ can potentially boost the rank of $H^{(0)}$, in the hope that the
$H^{(0)}$ reaches its maximal rank and Algorithm~\ref{alg:quasi} can correctly finds the minimal
realization.
\rvb{ However, for a given random process, the study of \cite{sontag1975some} showed that it is
  undecidable to verify whether it has a \emph{finite order } quasi-HMM realization. Even under our
  assumption that the process indeed has an \rvf{order $k$ minimal} quasi-HMM realization, it is still not
  clear how large the size of matrix $H^{(0)}$ ($d^n\times d^n$) needs to be so that it achieves the
  maximal rank $k$.
  In previous works, it was usually implicitly assumed that $N$ is large enough so that $H^{(0)}$
  achieves its maximal rank \cite{Balle2013spectral}.  Yet without a bound on $n$ or $N$ the
  computational complexity of the algorithm is ambiguous.}

\subsection{Generic analysis of information complexity }
We desire a small window size $N$ while guaranteeing the full column rank of the matrices $E$ and
$F$ defined in \eqref{eq:E-def} and \eqref{eq:F-def}.
The following theorem shows that if the random process is generated by an order $k$ HMM in general
position, then we only need window size $N > 4\lceil\log_d(k)\rceil +1$ to guarantee the correctness
of Algorithm~\ref{alg:quasi}.

\begin{theorem}
  [Window size $N$ for quasi-HMM]
  \label{thm:window-size}
  \
  \begin{enumerate}[label={(\arabic*)}, leftmargin=*]
  \item Consider $\Ta_{(d,k)}^h$, \rvf{the class of all HMMs with output alphabet size $d$ and
    order $k$.}
    There exists a measure zero set $\mc E\in \Theta_{(d,k)}^h$, such that for all the output
    process generated by HMMs in $\Ta_{(d,k)}^h\backslash \mc E$, Algorithm~\ref{alg:quasi}
    \rvc{returns a} minimal quasi-HMM realization, if window size $N= 2n+1$ for some $n$ such that:
    \begin{align}
      \label{eq:n-cond-window}
      n > 8\lceil\log_d(k)\rceil.
    \end{align}
  \item For any pair of $(d, k)$, randomly pick an instance from the class $\Ta_{(d,k)}^h$. If for a
    given window size $N=2n+1$, the matrix $H^{(0)}$ achieves its maximal rank $k$, then for all
    HMMs in $\Ta_{(d,k)}^h$, excluding a measure zero set, $N$ is sufficiently large for the
    correctness of Algorithm~\ref{alg:quasi}.
  \end{enumerate}
\end{theorem}
\rvb{Since the elements of matrices $E$ and $F$ are polynomials of the parameters $Q$ and $O$, in order
to show $E$ has full column rank for $Q$ and $O$ in general position, it suffices to construct an
instance of HMM for which the matrix $E$ has full column rank.  In particular, we fix the transition
matrix $Q$ and randomize the observation matrix $O$ and bound the singular values of $E$ in
probability. The detailed proof is provided in Appendix~\ref{sec:proofs}.}

For all $(d,k)$ pairs in the set $\{2\le d\le k<3000\}$, we implemented the test in Theorem
\ref{thm:window-size} (2), and found that for all these cases $n =\lceil \log_d(k) \rceil$ is
sufficient. We conjecture that in general, $n \ge \log_d(k)$ is enough.

\rvd{In the worst case \cite{vidyasagar2011complete}, the ``Hankel rank'' of the matrix
$H^{(0)}$ with infinite window size can be larger than the rank of any finite size block
of the infinite matrix.
Instead of the worst case analysis, our generic analysis examines the average cases, and
it has the following implications: if the process \rvf{is generated by some average case
  HMM of order $k$}, then the Hankel rank equals $k$; moreover, the window size $n$ only
needs to be in the order of $O(\log_d(k))$ so that the rank of finite matrix $H^{(0)}$
achieves the Hankel rank.  }

\subsection{Existence of hard cases}
We showed that for generic HMM output processes, Algorithm~\ref{alg:quasi} is has polynomial
runtime. There exists a long line of hardness results for learning HMMs
\cite{kearns1994learnability,mossel2005learning,terwijn2002learnability}, showing that in the worst
case (lie in the measure zero set in the parameter space) learning the distribution of an HMM {\em
  can} be computationally hard under cryptographic assumptions.

\begin{figure}[h!]
  \centering{
    \includegraphics[width = 0.47\textwidth]{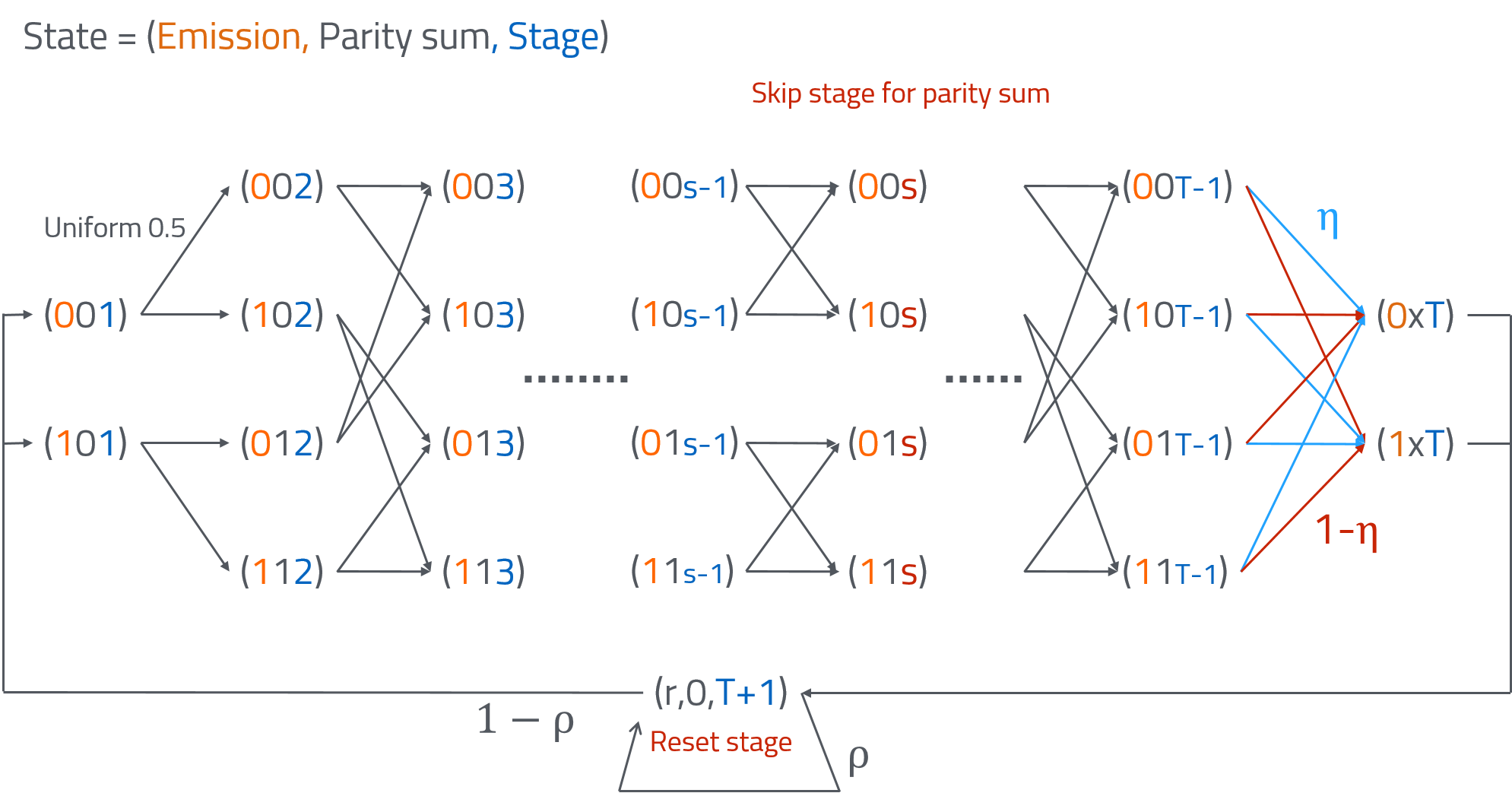}
  }
  \caption{Reduction of HMM to noisy parity to lower bound the worst case computational
    complexity.
    In the state transition diagram, for stage $t= 1,\cdots, T-1$, the emission state $E_{t}$ is
    uniformly distributed over $\{0,1\}$ and is observed.  For stage $t= 2,\cdots, T-1$, the parity
    state $S_{t}$ computes $E_{t-1}\oplus S_{t-1}$, except for at one unknown stage $s$, $S_{t} =
    S_{t-1}$.  At stage $T$, with probability $\eta$, the correct parity state $S_{T-1}$ is
    revealed, and with probability $1-\eta$, the complement is observed.  $(T+1)$ is a reset stage,
    with probability $\rho$ it stays in the reset stage.  }
  \label{fig:noisy-parity}
\end{figure}

In Fig.~\ref{fig:noisy-parity}, we adapt the hardness results to our setting and give an example.
The state diagram describes the transition and observation probabilities.
Solving the realization problem is equivalent to learning the joint distribution of the process. One
can verify that the window size $N$ needs to be at least as large as $T$, which is proportional to
the order of the underlying HMM, and therefore the computation complexity is exponential in the
order of the HMM.

\rva{ We point out that not all HMMs in the measure zero set are information theoretically hard to
  learn.  For instance, consider the degenerate HMM in
  \cite{allman2009identifiability} with the transition matrix $Q = I_{k\times k}$ and with general
  position observation matrix $O$.  Suppose that $d\ll k$, it was shown that the window size $N$
  needs to be in the order of $k^{1\over d}$ so that matrices $E$ and $F$ attain full column rank.
  However the distribution of this i.i.d. process is not fundamentally difficult to learn.
It remains an open problem to find realization algorithm that can handle more cases.
}

\subsection{Stability analysis}
In practice, the joint probabilities in $\mc P^{(N)}$ are estimated based on finite sample sequences
of the process.  In the next theorem, we show that in order to achieve $\epsilon$-accuracy in the
parameters of the minimal quasi-HMM realization, the number of sample sequences we need to estimate
$\mc P^{(N)}$ is polynomial in all relevant parameters, including the order $k$.

\begin{theorem}
  \label{thm:sample-quasi}
  Given $T$ independent sample sequences of the output process of an HMM of order $k$ and with
  alphabet size $d$. Construct $\wh H^{(0)}$ and $\wh H^{(j)}$'s as in \eqref{eq:H0-def} and
  \eqref{eq:Hj-def} with the empirical probabilities.
  Let $N = 2n + 1$, and $n = 2\lceil\log_d(k)\rceil$.
  Let $\wh \ta^o=(k,\wt u,\wt v,\wt A^{(j)}:j\in[d])$ and $\wt \ta^o=(k,\wt u,\wt v,\wt
  A^{(j)}:j\in[d])$ be the output of Algorithm~\ref{alg:quasi} with the empirical probabilities and
  the exact probabilities for the input, respectively.
  Then, in order to achieve $\epsilon$-accuracy in the output with probability at \rvc{least} $1-\eta$, namely:
  \begin{align*}
    \|\wh u -\wt u\|  \le \epsilon,\  \|\wh v -\wt v\|  \le \epsilon, \
    \|\wh A^{(j)} - \wt A^{(j)} \|\le \epsilon, \fa j,
  \end{align*}
  the number of independent sample sequences we need is given by:
  \begin{align*}
    T = {C k^6 d^4\over \epsilon^4 \sigma_k^8}\log\lt({ 2k^4 d^2\over \eta} \rt),
  \end{align*}
  where $\sigma_k$ is the $k$-th singular value of $ H^{(0)}$ and $C$ is some absolute constant.
\end{theorem}

Since the core of the algorithm is singular value decomposition of the matrix $H^{(0)}$, the
stability analysis mostly uses the standard matrix perturbation results, which we review in
Appendix~\ref{sec:auxiliary-lemmas}. The detailed proof is provided in Appendix~\ref{sec:proofs}.

\begin{remark}
  Note that Theorem~\ref{thm:window-size} shows that for window size $N$ large enough
  ($O(\log_d(k))$), the exact realization problem (no estimation noise) can be solved with poly time
  algorithm.
  When empirical probabilities are used, Theorem~\ref{thm:sample-quasi} shows that the
  required number of independent samples is polynomial in $k$, $d$, and $1/\sigma_k$. $\sigma_k$
  depends on the HMM that generates the process.
  In the proof of Theorem~\ref{thm:window-size}, it is showed that there exist cases for which
  $\sigma_k$ is lower bounded by constant, for which case the sample complexity is indeed
  polynomial; however there also exists hard cases for which $\sigma_k$ is arbitrarily small.
  We defer the analysis of sample complexity, which relies on understanding the relation between
  window size, HMM parameter, and $\sigma_k$, to future work.

\end{remark}


\section{Minimal HMM Realization Problem}
\label{sec:minim-hmm-real}
Recall that an HMM can be easily converted to a quasi-HMM of the same order as shown in
\eqref{eq:o-to-h-u}--\eqref{eq:o-to-h-A}, yet given a quasi-HMM realization it is difficult to
construct an HMM \cite{anderson1999realization}.
In this section, we apply tensor decomposition techniques to study the minimal HMM
realization problem and discuss its connection to the previous section.
In particular, we show that for processes generated by general position HMMs, the two
realization problems have similar computational complexity.

\subsection{Preliminaries on tensor algebra}
\paragraph{Definitions}
Tensor algebra has many similarities to but also many striking differences from matrix algebra, one
of which is that, under very mild conditions, \rvb{tensor \emph{minimal rank} decomposition is unique up
to column scaling and permutation, which is the key property  exploited to uniquely identify the
\emph{minimal} HMM realization.  This is in parallel with the fact that we use matrix rank
decomposition to find a minimal quasi-HMM realization.}

We review some properties of \rve{3rd} order tensors below. A more detailed introduction to tensor
algebra can be found in \cite{kolda2009tensor} and the references therein.
One way to view a \rve{3rd} order tensor $X\in \mbb R^{n_A\times n_B\times n_C}$ is that it defines a
three-way array,  multi-indexed by $(j_1,j_2,j_3)$, $\fa j_1\in[n_A],
j_2\in[n_B], j_3\in[n_C]$.
A rank-1 tensor $X= a\ot b\ot c$ is defined to be the outer-product of the three vectors
$a,b,c$ and $X_{j_1,j_2,j_3} = a_{j_1}b_{j_2}c_{j_3}$.
\rvb{Tensor rank decomposition} is a natural extension of matrix singular value
decomposition (SVD) to higher order tensors.

\begin{definition}[Tensor rank decomposition]
  The rank decomposition of a \rve{3rd} order tensor $X\in\mbb R^{n_A\times n_B\times n_C}$ is a sum of
  rank-1 tensors \rvb{for the smallest number of summands $k$:}
\begin{align*}
  &X = A \ot B\ot C = \sum_{i=1}^{k} A_{[:,i]}\ot B_{[:,i]} \ot C_{[:,i]},
\end{align*}
where matrices $A \in\mbb R^{n_A\times k}$, $B\in\mbb R^{n_B\times k}$, $C\in\mbb R^{n_C\times
  k}$. The minimal number of summands $k$ is defined to be the rank of the tensor.
\end{definition}

A tensor can also be viewed as a multi-linear operator.  Consider a \rve{3rd} order tensor $X$.  For given $m_A,m_B,m_C$, it
defines a multi-linear mapping $X(V_1,V_2,V_3): \mbb R^{m_A\times n_A}\times \mbb R^{m_B\times n_B}\times\mbb
R^{m_C\times n_C} \to \mbb R^{m_A\times m_B\times m_C}$ as below: $(\fa j_1\in[m_A], j_2\in[m_B], j_3\in[m_C])$
  \begin{align}
    \label{eq:tensor-def-1}
    &[X(V_1, V_2, V_3)]_{j_1,j_2,j_3} \\
    =& \sum_{ i_1\in[n_A], i_2\in[n_B], i_3\in[n_C]} X_{i_1,i_2,i_3} [V_1]_{j_1,i_1} [V_2]_{j_2,i_2}
    [V_3]_{j_3,i_3}.
    \nonumber
  \end{align}
  Assuming that the tensor admits a decomposition $X = A\ot B\ot C \in \mbb R^{n_A\times n_B\times
    n_C}$, \rvc{we can equivalently write:}
  \begin{align}
    \label{eq:tensor-def-2}
    X(V_1, V_2, V_3) = (V_1A)\ot (V_2B) \ot (V_3C),
  \end{align}
  Note that $X$ can have different forms of decompositions, yet the mappings defined in
  \eqref{eq:tensor-def-2} are all equivalent.

\begin{definition}[Khatri-Rao product]
  For matrices $A \in\mbb R^{n_A\times k}$, $B\in\mbb R^{n_B\times k}$, the (column) Khatri-Rao product $X= A\odot B
  \in\mbb R^{n_An_B\times k}$ is defined as follows:
  \begin{align*}
    X_{(j_1-1)n_B+j_2,i} = A_{j_1,i} B_{j_2,i},
  \end{align*}
  and each column of $X$ is a rank-1 Khatri-Rao product.
\end{definition}

An equivalent representation of a \rve{3rd} order tensor $ X\in\R^{n_A\times n_B\times
  n_C}$ is given by its matricization, obtained by rearranging the elements of the tensor
into a matrix.  For example, the matricization along the third mode gives a matrix $\ol
X^{(3)}$ is specified as below:
\begin{align*}
 \left[ \ol X^{(3)} \right]_{j_3,\ ((j_1-1)n_B+j_2)} = X_{j_1,j_2,j_3}.
\end{align*}
Moreover, if the tensor admits a decomposition $X = A\ot B\ot C$, we can write the
matricization as Khatri-Rao product of the factors: $\ol X^{(3)} = C(A\odot B)^\top$.

\paragraph{Uniqueness condition}
Unlike the rank decomposition of matrices, under rather mild conditions of the factors we
can uniquely (up to common column permutation and scaling) identify the factors from the
3rd order tensor $X$
In the following, we state a set of sufficient conditions on the factors $A,B,C$ that
guarantee the uniqueness of tensor decomposition,


\begin{definition}[Kruskal rank]
  The Kruskal rank of a matrix $A\in\mbb R^{n\times m}$ equals $r$ if any set of $r$ columns of $A$ are linearly
  independent, and there exists a set of $(r+1)$ columns that are linearly dependent (if $r<m$).
\end{definition}

\begin{lemma}[Uniqueness of tensor decomposition (\cite{kruskal1977three,sidiropoulos2000uniqueness})]
  \label{prop:suff-cond-id}
  The tensor factorization $X = A\ot B\ot C$ is unique up to column permutation and
  scaling, if
 \begin{align}
   \label{eq:k-rank-identifiability}
    krank(A) + krank(B) + krank(C) \ge 2k+2.
  \end{align}
\end{lemma}

\paragraph{Decomposition algorithms}

Unlike matrix SVD, in general tensor decomposition is a hard problem \cite{kolda2009tensor}.
Nevertheless, for cases where the factors $A, B, C$ satisfy certain rank conditions, there exist
efficient and provable algorithms. We include the detailed steps of the algorithm in
the appendix for completeness.

If the matrix $A$ and $B$ both have full column rank, Algorithms~\ref{alg:1} in the appendix
(\cite{leurgans1993decomposition}) can uniquely recover the factors up to common column permutation,
with running time polynomial in the dimension of the tensor.  Other algorithms such as tensor power
method and recursive projection, which are possibly more stable in practice, also apply here.

Algorithm~\ref{alg:2} is another efficient tensor decomposition algorithm
(\cite{de2007fourth} \cite{jiang2004kruskal}) to a subset of the degenerate instances
whose transition matrix is rank deficient.
Instead of requiring both $A$ and $B$ to be of full rank $k$, this algorithm requires that
the factor $C$ and the Khatri-Rao product $A\odot B$ have full column rank $k$.
The basic idea of the algorithm is as follows: there is a {\em unique rank decomposition}
of the \rve{3rd} dimension matricization of the tensor: $\ol X^{(3)}= F E^\top = C (A\odot
B)^\top$, under the algebraic constraints that each column of the matrix $E$ is a rank one
Katri-Rao product.

\subsection{Minimal HMM realization}

%
\paragraph{Formulation}
For a fixed window size $N=2n+1$, given the exact joint probabilities in $\mc P^{(N)}$, similar to
the construction of $H^{(0)}$ in \eqref{eq:H0-def}, one can construct a \rve{3rd} order tensor $M\in \mbb
R^{d^n\times d^n\times d}$ as below:
\begin{align}
  \label{eq:def-M}
  M_{\ L(\mb l_1^{n}), L(\mb l_{-1}^{-n}), \ l_{0}} = \mbb P\Big(\mb y_{-n}^{n} =\mb
  l_{-n}^{n}\Big), \quad \fa \mb l_{-n}^{n}\in\mbb [d]^{N}.
\end{align}
\rvb{Suppose that the process \rvf{has a  minimal HMM realization $\ta^h = (k,
  Q,O)$ of order $k$}.}  We can write $M$ as a tensor product:
\begin{align}
  M = A\ot B\ot C,
  \label{eq:M-ABC}
\end{align}
where the matrices $A,B\in\mbb R^{d^n\times k}$ and $C\in\mbb R^{d\times k}$ correspond to the
conditional probabilities:
\begin{align}
  A_{L(\mb l_1^{n}), m} &= \mbb P\Big(\mb y_1^{n} =\mb l_1^{n} \Big| x_0=m\Big),
  \label{eq:A-to-P}
  \\
  B_{L(\mb l_{-1}^{-n}), m} &= \mbb P\Big(\mb y_{-1}^{-n} =\mb l_{-1}^{-n} \Big| x_0=m\Big),
  \label{eq:B-to-P}
  \\
  C_{l,m} & = \mbb P\Big(y_0 = l, x_0= m \Big).
  \label{eq:C-to-P}
\end{align}
%
Moreover, observe that $A$ anh $B$ are recursive linear functions of the model parameters $Q$ and
$O$ as below:
\begin{align}
  \label{eq:ABC-def1}
  A^{(n)} & = \mbb P\Big(\mb y_1^{n} \Big| x_0=m\Big) = (O\odot A^{(n-1)})Q,
  \\
  \label{eq:ABC-def2}
  B^{(n)} &= \mbb P\Big(\mb y_{-1}^{-n}\Big| x_0=m\Big) = (O\odot B^{(n-1)})\wt Q,
\end{align}
and $A^{(1)} = OQ$ and $B^{(1)} = O\wt Q$. In particular, for the given window size $N = 2n+1$, we have:
\begin{align}
  \label{eq:ABC-def3}
  A = A^{(n)}, \quad B = B^{(n)}, \quad   C = O Diag(\pi).
\end{align}
The basic idea of \emph{recovering} the minimal HMM realization $\ta^h$ (up to hidden state
relabeling) is to first \emph{recover} the factors $A, B$ and $C$ via tensor decomposition, and then
extract the transition and observation probabilities from the factors.
\rvb{ The minimal order condition is again reflected in the tensor \emph{rank}
  factorization, as any HMM realization of lower order results in a tensor $M$ of lower
  tensor rank, which is a contradiction.}

\paragraph{Identifiability}
The identifiability of the minimal HMM relies on the fact that the tensor rank
decomposition indeed recovers the factor $A,B,C$ defined in
\eqref{eq:A-to-P}--\eqref{eq:C-to-P}.
Note that by definition, the column stochastic observation matrix $O$ must have Kruskal
rank greater than $2$, otherwise there exist two identical columns in $O$, and the
corresponding two hidden states can be merged to give an equivalent HMM realization of
smaller order.

\begin{lemma}[Uniqueness of tensor decomposition]
  \label{cor:suff-id-map1}
  Given window size $N$, if the matrices $A,B \in\mbb R^{d^n\times k}$ defined in
  \eqref{eq:ABC-def1}--\eqref{eq:ABC-def3} have full column rank $k$, then $M$ can be
  uniquely decomposed into column stochastic matrices $A, B, C$ as in \eqref{eq:M-ABC} (up
  to common column permutation).
\end{lemma}

In parallel with Theorem~\ref{thm:window-size}, the next theorem shows that the condition
above is satisfied for a general position HMM process  with sufficiently large window size
$N$.
\begin{theorem}[Choice of $N$ for HMM realization]
  \label{thm:info-compl-hmm}
  Consider $\Ta^{h}_{(d,k)}$, the \rvf{class of all HMMs with output alphabet size $d$ and order $k$}.
  There exists a measure zero set $\mc E \in \Theta_{(d,k)}^h$ such that for all output processes
  generated by HMMs in the set $\Theta_{(d,k)}^h\backslash \mc E$, the minimal quasi-HMM realization
  can be computed based on the joint probabilities in $\mc P^{(N)}$, if window size $N= 2n+1$ for
  some $n$ such that:
  \begin{align}
    \label{eq:pick-n-svd}
    n > 8\lceil\log_d(k)\rceil.
  \end{align}
\end{theorem}

\paragraph{ Algorithms}
\label{sec:effic-algor-minim}

\rvb{The matrices $A,B$ and $C$, defined in \eqref{eq:ABC-def1}--\eqref{eq:ABC-def3}, are
  polynomial functions of the parameters $Q$ and $O$ of the minimal HMM realization.} The
following theorem exploits the recursive structure of these polynomials to recover the
parameters $Q$ and
$O$ if the factors $A,B,C$ are given.

\begin{theorem}[Recovering $Q$ and $O$ from $A,B,C$]
  \label{thm:recover-QO-from-ABC}
  Given the matrix $C$, one can obtain the observation matrix by:
  \begin{align}
    O_{[:,i]} = {C_{[:,i]} /( \mb e^\top C_{[:,i]})}, \quad \fa i\in[k].
    \label{eq:find-O}
  \end{align}
  Given the matrix $A\in\R^{d^n\times k}$, we first scale each of the column similar to \eqref{eq:find-O} so that each
  column is stochastic, and corresponds to the conditional probabilities $\mbb P(\mb y_1^n|x_0)$ as
  shown in \eqref{eq:A-to-P}. We marginalize the conditional distribution to get $ A^{(1)} = \mbb
  P\left(y_{1} |x_0 \right)\in\mbb R^{d\times k}$ and $ A^{(n-1)} = \mbb P\left(\mb y_1^{n-1} \big|
    x_0 \right) \in \mbb R^{d^{n-1}\times k}$.

  \begin{enumerate}[label={(\arabic*)}, leftmargin=*]
  \item If  $A$ has full column rank $k$ \rva{(\cite{allman2009identifiability})}:
    \begin{align}
      \label{eq:recover-QO-1}
      Q = \Big( O\odot A^{(n-1)}\Big)^{\dagger} A.
    \end{align}

  \item If $C$ has full column rank $k$:
    \begin{align}
      \label{eq:recover-QO-2}
      Q = O^{\dagger} A^{(1)}.
    \end{align}
    where $(X)^{\dagger} = (X^\top X)^{-1} X^\top$ denotes the pseudo-inverse of a matrix $X$.

  \end{enumerate}
\end{theorem}

In the proof of Theorem~\ref{thm:info-compl-hmm}, we show that for general position HMMs
with sufficiently large window size, the matrices $A$ and $B$ achieve full column rank
$k$.  When this holds, Algorithm~\ref{alg:1} computes the unique tensor decomposition to
recover the factors $A,B,C$.
Theorem \ref{thm:recover-QO-from-ABC} (1) applies to recover $Q$ and $O$ from the factors.

However, if the transition matrix $Q$ of the minimal HMM realization does not have full
rank, and no matter how large the window size is, the matrix $A$ never achieves full
rank. Note that these HMMs are degenerate cases belonging to the measure zero set in
Theorem~\ref{thm:info-compl-hmm}, and Algorithm~\ref{alg:1} is not applicable for
decomposing the tensor $M$.
However, it is still possible to apply Algorithm~\ref{alg:2}.  Note that a necessary
condition for it to work is that $d\ge k$ and the observation matrix is of full column
rank.

\rva{Let $\Theta_{(d,k,r)}^h$ denote the model class of HMMs with output alphabet $d$ and
  order $k$, for $d\ge k$ and the transition matrix $Q$ has rank $r<k$. Note that
  $\Theta_{(d,k,r)}^h$ is a subset of the measure zero set $\mc E$ in
  Theorem~\ref{thm:info-compl-hmm}.
  The following theorem shows that if Algorithm~\ref{alg:2} runs correctly for a random
  instance in this subset, then the algorithm works for almost all HMMs in this subset.}

\begin{theorem}[Correctness of Algorithm~\ref{alg:2}]
  \label{thm:correct-alg2}
  Given $d,k$ and $r$ and consider the set $\Theta_{(d,k,r)}^h$. Let $A,B,C$ be defined as in
  \eqref{eq:ABC-def1}-- \eqref{eq:ABC-def3} for $n=1$, and let $M=A\ot B\ot C$.
  If Algorithm~\ref{alg:random-check} returns ``yes'', then there exists a measure zero set $\mc E
  \in \Theta_{(d,k,r)}^h$, such that Algorithm~\ref{alg:2} returns the tensor decomposition $M =A\ot
  B\ot C$ for all HMMs in the set $\Theta_{(d,k,r)}^h\backslash \mc E$.
  Moreover, if the latter is true, Algorithm \ref{alg:random-check} returns ``yes'' with
  probability 1.
\end{theorem}

For this class of degenerate HMMs, Theorem~\ref{thm:recover-QO-from-ABC} (2) applies to
recover $Q$ and $O$.

Note that for both the general position case and this degenerate case, the computation
complexity to recover the parameters of the minimal HMM realization are polynomial in both
$d$ and $k$, and this is an immediate result of the log upper bound of the window size.

\begin{algorithm}
  \caption{Check Condition}
  \label{alg:random-check}
  \begin{enumerate}
  \item Randomly choose an HMM from  $\ta^h\in\Ta_{(d,k,r)}^h$.
  \item Construct matrices $A,B,C$ with $(Q,O)$ as defined in
    \eqref{eq:ABC-def1}--\eqref{eq:ABC-def3} for $n=1$, namely $ A = OQ$, $B = O\wt Q$, and $ C =
    ODiag(\pi)$.
  \item Let $M = A\ot B\ot C$.  Run Algorithm \ref{alg:2} with the input $M$.
  \item Return ``yes'' if the algorithm returns $A,B,C$ uniquely up to a common column permutation,
    and ``no'' otherwise.
  \end{enumerate}
\end{algorithm}

\section{Conclusion}
In this paper, we discussed two realization problems.  We show that for output processes generated
by HMMs in general position, both learning the minimal quasi-HMM realization and learning the real
minimal HMM realization are easy-- in the sense that there exist efficient algorithms to compute the
minimal realizations with running time and sample complexity both polynomial in the relevant
parameters of the problem.

\bibliographystyle{plain}
\bibliography{hmm}

\appendices


\rvd{\section{Tensor decomposition algorithms}}

For completeness, we list two standard tensor decomposition algorithms in this section.

\begin{algorithm}[h!]
  \caption{Simultaneous diagonalization for \rve{3rd} order tensor decomposition
    \cite{leurgans1993decomposition} }
  \label{alg:1}
  \begin{algorithmic}
    \REQUIRE { A \rve{3rd} order tensor $M\in\mbb R^{d^{n}\times d^{n}\times d}$ }
    \ENSURE { $k,  A,  B\in\R^{d^n\times k }, C\in\R^{d\times k}$}
    \begin{enumerate}[leftmargin=*]
      \STATE Randomly pick two unit norm vectors $\mb v_1, \mb v_2\in \mbb R^{d}$. Project $M$ along
      the \rve{3rd} dimension to obtain two matrices:
      \begin{align*}
        \wt M_{1} = M(I,I, \mb v_1),\quad \wt M_{2} = M( I, I, \mb v_2).
    \end{align*}

    \STATE Compute the eigen-decomposition of matrix $(\wt M_1 \wt M_2 ^{-1})$ and $(\wt M_2 \wt M_1
    ^{-1}) $, and  let the columns of matrix $A$ and $B$ be the eigenvectors of $(\wt M_1 \wt M_2 ^{-1})$
    and $(\wt M_2 \wt M_1 ^{-1})$, respectively.

    Scale the columns of $A$ and $B$ to be stochastic, and pair the eigenvectors in $A$ and $B$
    corresponding to the reciprocal eigenvalues, namely:
    \begin{align*}
      \wt M_1 \wt M_2 ^{-1} =  A \Lambda  A^{-1}, \quad \wt M_2 \wt M_1 ^{-1} =
       B \Lambda^{-1}  B^{-1}.
    \end{align*}

    \STATE Let $k$ be the number of non-zero eigenvalues.

    \STATE Let $\ol M^{(3)}\in\mbb R^{d^{2n}\times d}$ be the \rve{3rd} dimension matricization of
    $M$.  Set $C$ to be:
    \begin{align*}
      C = \ol M^{(3)} ( (A\odot B)^\dag)^\top
    \end{align*}

    \end{enumerate}

\end{algorithmic}
\end{algorithm}

\begin{algorithm}[h!]
  \caption{FOOBI for \rve{3rd} order tensor decomposition}
  \label{alg:2}
  \begin{algorithmic}
    \REQUIRE $M\in\mbb R^{d\times d\times d}$

    \ENSURE { $k, A, B, C.$}
    \begin{enumerate}[leftmargin=*]
      \STATE Let $\ol M^{(3)}$ be the \rve{3rd} dimension matricization of $M$.  Compute its SVD $\ol
      M^{(3)}=V_HD_HU_H^\top$.

      \STATE Set $k$ to be the number of non-zero singular values.  Let $F = V_HD_H^{1/2}$, and $E =
      U_HD_H^{1/2}$.

      \STATE Construct matrices $\{E^{(r)}\in\mbb R^{d\times d} : r\in[k]\}$:
    \begin{align*}
      [E^{(r)}]_{i,j} = E_{(i-1)d+j, r}, \fa i,j\in[d], \ \fa r\in[k].
    \end{align*}
    Construct the 4-th order tensors $\{P^{(r,s)}\in\mbb R^{d\times d\times d\times d}:
    r,s\in[k]\}$:
    \begin{align*}
      &[P^{(r,s)}]_{i_1,i_2,j_1,j_2}
      \\
      &=[ E^{(r)}]_{i_1,j_1}[E^{(s)}]_{i_2,j_2} + [E^{(s)}]_{i_1,j_1}[E^{(r)}]_{i_2,j_2}
      \\
      &-[E^{(r)}]_{i_1,j_2}[E^{(s)}]_{i_2,j_1} -[E^{(s)}]_{i_1,j_2}[E^{(r)}]_{i_2,j_1}.
    \end{align*}

    \STATE

    Compute a basis $\{H^{(i)}:i\in[k]\}$ of the $k$-dimensional kernel of
    $\{P^{(r,s)}:r,s\in[k]\}$:
    \begin{align*}
      \sum_{r,s=1}^{k} H_{r,s}^{(i)}P^{(r,s)} = 0,\ \tx{ s.t. }
      H_{r,s}^{(i)} = H_{s,r}^{(i)},  \fa r,s\in[k].
    \end{align*}

    \STATE Find the unique $W\in\mbb R^{k\times k}$ that simultaneously
    diagonalizes the basis:
    \begin{align*}
      H^{(i)} = W \Lambda^{(i)} W^\top, \ \fa i\in[k].
    \end{align*}

    \STATE Let $C = F (W^{-1})^\top$ and $A\odot B = EW$.  Compute the rank one decomposition of
    each column of $A\odot B$, with proper normalization such that $A$ and $B$ are column
    stochastic.

    \end{enumerate}

\end{algorithmic}
\end{algorithm}

\section{Proofs}
\label{sec:proofs}
\tb{\noindent(Proof of Lemma~\ref{prop:correct-quasi-alg})}


If both $E$ and $F$ have full column rank $k$, by Sylvester inequality the rank of the matrix
$H^{(0)}$ is also equal to $k$, the order of minimal quasi-HMM realization.
  %
Therefore, for the two matrices $U$ and $V$ obtained in Step 2 in Algorithm~\ref{alg:1}, there
exists some full rank matrix $W\in \mbb R^{k\times k}$ such that:
\begin{align*}
  U = EW, \quad V^\top = W^{-1} F^\top.
\end{align*}
Therefore, Step 3 returns
\begin{align*}
  \wt A^{(j)} = W^{-1 }E^{\dagger} E A^{(j)} F^\top (F^\top)^{\dagger } W = W^{-1} A^{(j)} W.
\end{align*}
By the normalization constraint in Definition~\ref{def:quasi-hmm}, we have
\begin{align*}
  u^\top W = u^\top \sum_{j=1}^{d} A^{(j)} W = u^\top W \sum_{j=1}^{d}\wt A^{(j)}.
\end{align*}
Moreover, since
  \begin{align*}
   U  &= \left[
    \begin{array}[c]{c}
      u^\top (A^{(1)}\cdots A^{(1)})
      \\
      u^\top (A^{(1)}\cdots A^{(2)})
      \\
      \vdots
      \\
      u^\top(A^{(d)}\cdots A^{(d)})
    \end{array}
  \right]W
  = u^\top W\left[
    \begin{array}[c]{c}
     \wt A^{(1)}\cdots \wt A^{(1)}
      \\
      \wt A^{(1)}\cdots \wt A^{(2)}
      \\
      \vdots
      \\
      \wt A^{(d)}\cdots \wt A^{(d)}
    \end{array}
  \right],
  \end{align*}
  in Step 2 we obtain $ \wt u^\top = u^\top W,$ and similarly, we can argue that $ \wt v = W^{-1} v$.
  Thus we conclude that the output $\wt \ta^o=(k,\wt u,\wt v, \wt A^{(j)}:j\in[d])$ is a valid
  minimal quasi-HMM realization of order $k$, and is equivalent to $\ta^o$ up to a linear
  transformation.

  \qed

\tb{\noindent(Proof of Theorem~\ref{thm:window-size})}

Assume that the observed process \rvf{has  a minimal  HMM realization $\ta^h$ of order $k$}, i.e.,
 $\ta^h\in\Ta^h_{(d,k)}$,
and let $\ta^o$ denote the equivalent order $k$ quasi-HMM as shown in
\eqref{eq:o-to-h-u}-\eqref{eq:o-to-h-A}.  For window size $N=2n+1$, define the matrices $E$ and $F$
for $\ta^o$ as in \eqref{eq:E-def} and \eqref{eq:F-def} and note that:
  \begin{align*}
    &E_{L(\mb l_1^n),i} \\& = [ u^\top  (A^{(l_n)}\cdots A^{(l_1)})]_i
    \\
    & = \mb e^\top  \mbb P(x_{n},y_{n-1}=l_n |x_{n-1}) \cdots \mbb P(x_{1},y_{0}=l_1 |x_{0} = i)
    \\
    &= \mbb P\Big (\mb y_0^{n-1} = \mb l_1^n \Big|x_0 = i\Big),
  \end{align*}
  and similarly,
\begin{align*}
    F_{L(\mb l_1^n),i} & = [ A^{(l_n)}\cdots A^{(l_1)} \pi]_i
    = \mbb P\Big(\mb y_{-1}^{-n} =\mb l_1^n, x_0 = i\Big).
  \end{align*}

  Lemma~\ref{prop:correct-quasi-alg} shows that a sufficient condition for the correctness of
  Algorithm~\ref{alg:quasi} is that both $E$ and $F$ have full column rank $k$.
  In this proof, we show that when $Q$ and $O$ of the HMM $\ta^h\in\Ta^h_{(d,k)}$ are in general
  position, this rank condition is satisfied if the window size $N=2n+1$ satisfies
  \eqref{eq:n-cond-window}.

  \rvc{
  Note that the minors of $E$ and $F$ are polynomials in the elements of $Q$ and $O$, thus it
  defines a algebraic set in the parameter space by setting all the minors to zero to make $E$ and
  $F$ to be rank deficient.
  By basic algebraic geometry \cite{griffiths2014principles}, the algebraic set either occupies the
  entire Zariski closure or is a low-dimensional manifold of Lebesgue measure zero.  In particular,
  the Zariski closure of $\Ta_{(d,k)}^h$, defined to be the smallest algebraic set containing
  $\Ta_{(d,k)}^h$, is given by $\ol\Ta_{(d,k)}^h:=\{O\in\R^{d\times k}, Q\in\R^{k\times k}:\mb
  e^\top O = \mb e^\top, \mb e^\top Q = \mb e^\top\}$ (note that the element-wise non-negativity
  constraints can be omitted when considering the Zariski closure).
  Therefore, it is enough to show that for some specific choice of $Q$ and $O$ in
  $\ol\Ta_{(d,k)}^h$, the matrices $E$ and $F$ achieve full column rank $k$.
  Moreover to construct an instance, we can further ignore the stochastic constraints, as scaling
  does not the independence property of the columns in $E$ and $F$.}

  We fix the transition matrix $Q$ to be the state shifting matrix as below:
  \begin{align}
    Q_{i-1, i} = 1, \tx{ for } 2\le i\le k, \tx{ and } \quad Q_{k, 1} = 1,
  \end{align}
  Note that with this choice of $Q$, $\pi = {1\over k} \mb e$, and $\wt Q = Q^\top $. Due to the
  symmetry of the forward and backward transitions, we can focus on showing that $E$ has full column
  rank and the same argument applies to $F$.

  We randomize the observation matrix $O$ and let the columns be independent random variables
  uniformly distributed on the $d$-dimensional sphere.
  In order to show that there exists a construction of $(Q,O)$ such that $E$ has full column rank,
  it suffices to show that $E$ achieves full column rank with positive probability over the
  randomness of $O$.
  We apply Gershgorin's theorem to prove that the columns of $E$ are incoherent.

  Note that for the shifting matrix $Q$, we have:
  \begin{align*}
    E_{[:,i]}= O_{[:,i]}\odot \cdots O_{[:,i+n-1]}.
  \end{align*}
  Since we have $d\ge 2$ and $n<k$, for notational convenience, we slightly abuse notation to write
  the $j$-th column of $O$ as $O_{[:,j]}$, while for $k<j\le 2k$, it actually refer to the $(j-k)$-th
  column of $O$.

  Define matrix $X\in\R^{k\times k}$ to be:
  \begin{align*}
    X_{i,j} = E_{[:,i]}^\top  E_{[:,j]} = \prod_{m=0}^{n-1} (O_{[:,i+m]}^\top  O_{[:,j+m]}), \  \fa i,j\in[k].
  \end{align*}
  By the assumption that the columns of $O$ are uniformly distributed on the $d$-dimensional sphere,
  we have $ X_{i,i} = 1$, for all $ i\in[k]$.

  \rva{Fix some $\beta, \gamma = \beta^2 \in(0,1)$}. Suppose that, for any $i\neq j$,
  \begin{align}
    \label{eq:show-Xij-small}
    \mbb P\left(
      \left|X_{i,j} \right| <{\beta\over k}
    \right) > 1-{\gamma\over k^2}.
  \end{align}
  Then apply union bound on $j$, we have  for any $i$:
  \begin{align*}
    \mbb P\left(
      \sum_{j\neq i}^{k} \left|X_{i,j} \right| < \beta
    \right) &\ge
    \mbb P\left(
      \fa j\in[k],j\neq i, \left|X_{i,j} \right| <{\beta\over k}
    \right)
    \\
    &>
    1- {\gamma\over k}.
  \end{align*}
  Again apply union bound on $i$, we have:
  \begin{align*}
    \mbb P\left( \fa i\in[k], |X_{i,i}|-\sum_{ j\neq i} \left|X_{i,j} \right| \ge 1-\beta \right) >&
    1- k{\gamma\over k} =1-\gamma.
  \end{align*}
  Apply Gershgorin's theorem, we have that \rva{ with probability at least $\gamma$, the matrix $X=
    E^\top E$ is of full rank $k$, and the smallest singular value is at least $1-\beta$.}  There
  must exist some instance of $O$ such that this statement holds.

  Next, we verify the statement in \eqref{eq:show-Xij-small}.  Equivalently, we want to show that
  for $i\neq j$:
  \begin{align*}
    1- {\gamma\over k^2} &<
    \mbb P\left(
      \prod_{m=0}^{n-1} \left| O_{[:,i+m]}^\top  O_{[:,j+m]}\right| < {\beta\over k}
    \right)
    \\
    &=
    \mbb P\left(
      \sum_{m=0}^{n-1} \log( \left| O_{[:,i+m]}^\top  O_{[:,j+m]}\right|) < -\log( {k\over \beta})
    \right)
    \\
    &=\mbb P\left(
      \sum_{m=0}^{n-1} \log\left( { 1\over \left| O_{[:,i+m]}^\top  O_{[:,j+m]}\right|}\right)
      >\log( {k\over \beta})
    \right)
    \\
    & = \mbb P\left(
      \sum_{m=1}^{n} \log\left( { 1\over \left| v_{m} \right|}\right) >\log( {k\over \beta})
    \right)
  \end{align*}
  where $v_{m}$ are i.i.d. random variables with the distribution as the projection of a uniform
  unit-norm vector in $\R^{d}$ onto the first dimension.  The last equality is due to the
  independence of the  columns of $O$.

  Define the indicator random variable $s_{m}$ for $m\in[n]$:
  \begin{align*}
    s_m = \mb 1\left[\log({1\over |v_m|}) < {1\over c} \log(d)\right]
    = \mb 1\left[ |v_m| > {1\over d^{1\over c}} \right],
  \end{align*}
  where we pick constant $c=4$.
  Assume that $d\ge 2+( 8e)^2$ (as we really only care about the scaling), apply Johnson
  Lindenstrauss lemma (Lemma \ref{lem:sphere-proj}), setting $u_1$ to be $v_m$ and $t$ to be $ {1 /
    d^{1\over c}}$, we have:
  \begin{align*}
    \mu &= \mbb P(s_m = 1)
    < {4\over\sqrt{d-2}} e^{-{d-2\over 2 d^{2/c}}} <  {1\over 2e}e^{-{d-2\over 2 d^{2/c}}}
  \end{align*}

Note that  by definition:
  \begin{align*}
    \sum_{m=1}^{n} \log\left( { 1\over \left| v_{m} \right|}\right) > \sum_{m=1}^{n} {1\over c}\log(d)(1-s_m).
  \end{align*}
  Therefore it suffices to show that
  \begin{align*}
    1- {\gamma\over k^2} &< \mbb P\left( \sum_{m=1}^{n} {1\over c}\log(d)(1-s_m) >\log( {k\over
        \beta}) \right),
  \end{align*}
  or equivalently,
  \begin{align*}
    { \gamma\over k^2}&>
    \mbb P\left( \sum_{m=1}^{n}s_m > n- c{\log( k/\beta)\over\log(d)} \right)
    \\&=
    \mbb P\left( \sum_{m=1}^{n}s_m > \alpha c{\log( k/\beta)\over\log(d)} \right)
  \end{align*}
  where we set $n = (1+\alpha) c \log_d(k/\beta)$ for some $\alpha>1$.

  Apply the multiplicative Chernoff bound (Lemma \ref{lem:chernoff}), by setting $X_m = s_m$ for $m
  = 1,\cdots n$, and set $ \delta n \mu = \alpha c{\log( k/\beta)\over\log(d)}$, and $ {e\over
    \delta} = {e n \mu \over \alpha c{\log( k/\beta)\over\log(d)}} = {1+ \alpha \over \alpha}e \mu
  <e^{- \sqrt{d}/2}<1,$
  then we have
  \begin{align*}
    \mbb P\left( \sum_{m=1}^{n}s_m > \alpha c{\log( k/\beta)\over\log(d)} \right)
    < \left( {1+\alpha\over \alpha}e\mu \right)^{ \alpha c{\log(k/\beta)\over\log(d)}}.
  \end{align*}
  We want to show that the RHS is less than $\gamma/k^2$.  Taking log, this is equivalent to:
  \begin{align*}
    \alpha c{\log(k/\beta)\over\log(d)} \log_d\left({\alpha\over (1+\alpha )e\mu}\right) > {
      \log(k^2/\gamma) \over \log(d)}
  \end{align*}

  Recall that we have $\gamma = \beta^2$, ${1+\alpha\over \alpha}e \mu \le e^{-{d-2\over 2
      d^{2/c}}}$, $c= 4$ the above inequality holds if we pick $\alpha = 4/c = 1$, as
  \begin{align*}
    \alpha c \log_d({\alpha\over (1+\alpha )e\mu}) \ge 4{\log (e^{\sqrt{d}\over 2}) \over \log(d)}
    \ge 2 {\sqrt{d}\over \log(d)} \ge 2.
  \end{align*}
  Now we can conclude that \eqref{eq:show-Xij-small} holds.

\qed

  \tb{\noindent(Proof of Theorem~\ref{thm:sample-quasi})}

  Recall that the output of Algorithm~\ref{alg:quasi} is given by:
  \begin{align*}
    & \wh A^{(j)} = \wh D^{-1/2} \wh U_H^\top \wh H^{(j)}\wh V_H \wh D^{1/2},
    \\
    & \wh u = \wh D^{-1/2} \wh U_H^\top \mb e, \quad \wh v = \wh D^{-1/2} \wh V_H^\top \mb e,
  \end{align*}
  where $\wh U_H$ and $\wh V_H$  are the first $k$ left and right singular vectors of  $\wh H^{(0)}$, and the diagonal
  matrix  $\wh D$  has the first $k$ singular values of $\wh H^{(0)}$ on its main diagonal.
  In order to bound the distance between $\wh A^{(j)}$ and $\wt A^{(j)}$, $\wh u$ and $\wt u$, $\wh v$ and $\wt v$, we
  analyze the perturbation bound for each of the factor separately and apply Lemma~\ref{lem:prod-perturb} to bound the
  overall perturbation of the product form.

  First, denote $E_j = \wh H^{(j)} - H^{(j)}$ for $j=0,1,\dots, d$. For any element in $E_j$ we can be bound its norm
  using Hoeffding's inequality (Lemma~\ref{lem:hoeffding} ): with probability at least $1-2e^{-2T\delta^2}$, the
  $(i_1,i_2)$-th element of $E_j$ is bounded by: $ \|[E_j]_{i_1,i_2}\| \le \delta\qq{<1}$.
  Moreover, apply union bound to $j$ and all elements in each $E_j$, with probability at least $1-2
  k^4d^3 e^{-2T\delta^2}$, for all $j = 0,1,\dots,d$, we have
  \begin{align*}
    \|E_j\|_F \le \sqrt{k d^n} \delta < k^{1.5} d^{0.5} \delta,
  \end{align*}
  where the last inequality is due to $d^n < k^2 d$.

  Second, we apply the matrix perturbation bound (Lemma~\ref{lem:wedin}) to bound the distance of the singular vectors:
  \begin{align*}
    \|\wh U_H - U_H\| \le {\sqrt{2} \|E_0\|_F \over \sigma_k(H^{(0)})},
    \quad
        \|\wh V_H - V_H\| \le {\sqrt{2} \|E_0\|_F \over \sigma_k(H^{(0)})}.
  \end{align*}
  And we can apply Mirsky's theorem (Lemma~\ref{lem:mirsky} ) to bound the distance of the singular values:
  \begin{align*}
    \|\wh D- D\| \le \|E_0\|_F.
  \end{align*}
  Denote $\Delta_i= \sigma_i(\wh H^{(0)}) - \sigma_i( H^{(0)})$ and let $\sigma_i = \sigma_i(H^{(0)})$.  Note that if
  $\|E_0\|\le \sigma_k / 2$, we have that for any $i=1,\dots k$, $|\Delta_i| \le \|E_0\| \le \sigma_i/2$, then
  \begin{align*}
     ( {1\over \sqrt{\sigma_i}}- {1\over \sqrt{\sigma_i + \Delta_i}})^2
    =& {1\over \sigma_i + \Delta_i } (\sqrt{1+\Delta_i/\sigma_i} - 1)^2
    \\
    \le & {2\over \sigma_i} ( \Delta_i/\sigma_i+ 2 -2 \sqrt{1 + \Delta_i/\sigma_i})
    \\
    \le & {2\over \sigma_i}(3 |\Delta_i| / \sigma_i )
    \\
    \le & {6\over \sigma_k^2}|\Delta_i|,
  \end{align*}
  where the first inequality is due to $|\Delta_i|\le \delta_i/2$, and the second inequality is due to
  $\sqrt{1 + \Delta_i/\sigma_i} \ge 1- |\Delta_i/\sigma_i|$.
  Therefore we have that
  \begin{align*}
    \| \wh D^{-1/2} - D^{-1/2}\| \le {\sqrt{6 \sum_{i=1}^{k} |\Delta_i| }\over \sigma_k}
    \le  {\sqrt{6 \sqrt{k}\|\wh D- D\|} \over \sigma_k}.
  \end{align*}

  Finally, we apply Lemma~\ref{lem:prod-perturb} to bound the output perturbation.  Note that $\|D^{-1/2}\| =
  1/\sqrt{\sigma_k}$, $\|U_H\| = 1, \|V_H\| =1$. Moreover note that the probabilities in each row of $H^{(j)}$ sum up to
  less than 1, therefore by Perron-Frobenius theorem we have  $\|H^{(j)}\| \le 1 $.
  Therefore we have
  \begin{align*}
    &\|\wh A^{(j)} - \wt A^{(j)} \|
    \\\le & 2^4 \lt({2\sqrt{6 k^{1/2}\|E_0\|_F}\over \sigma_k^{1.5}} +{2\sqrt{2}
      \|E_0\|_F\over \sigma_k^{2}} + {\|E_j\|\over\sigma_k}\rt)
    \\
    \le &2^4\lt( {2\sqrt{6} k^{0.75} d^{0.25} \delta^{0.5} \over \sigma_k^{1.5}} + {2\sqrt{2}k d^{0.5} \delta \over
      \sigma_k^2} + {k d^{0.5} \delta \over \sigma_k} \rt)
    \\
    \le & {144 kd^{0.5}\over \sigma_k^2}\delta^{0.5},
  \end{align*}
  where the first inequality is due to $\|E_j\|\le \|E_j\|_F$, and the second inequality is due to $\delta<1$ and
  $\sigma_k\le \sigma_1 \le 1$.

  Similarly we can bound $\|\wh u-\wt u\|$ and  $\|\wh v-\wt v\|$ by:
  \begin{align*}
    \|\wh u -\wt u\| &\le \|\wh D^{-1/2} \wh U_H^\top - D^{-1/2} U_H^\top\| \sqrt{d^n} \le {4 k^{1.5} d \over
      \sigma_k^{1.5} }\delta^{0.5}.
  \end{align*}

  In summary, if we want to achieve $\epsilon$ accuracy in the output.
  we need $\delta$ to be no larger than $ \epsilon^2\sigma_k^4 /(144k^3 d^2)$. Set the failure probability to be $\eta =
  2k^4 d^3e^{-2T\delta^2}$, then number of sample sequences needed to estimate the empirical probabilities is given by:
  \begin{align*}
    T = 2 {144^2 k^6 d^4\over \epsilon^4 \sigma_k^8}\log\lt({ 2k^4 d^3\over \eta}
    \rt).
  \end{align*}

\qed

\tb{\noindent (Proof of Theorem~\ref{thm:info-compl-hmm})}

With exactly the same argument and constructional proof as for Theorem~\ref{thm:window-size}, we can
show that for the window size $N=2n+1$ satisfies \eqref{eq:pick-n-svd}, the matrices $A$ and $B$
have full column rank. By Lemma~\ref{cor:suff-id-map1} we have that the tensor decomposition of $M$
is unique.  Moreover, by the argument in Theorem~\ref{thm:recover-QO-from-ABC} (1), we have that the
model parameters $Q,O$ can be uniquely recovered from the factors $A,B,C$.  Thus in conclusion $\mc
P^{(N)}$ is sufficient for finding the minimal HMM realization.

\qed

\tb{\noindent  (Proof of Theorem~\ref{thm:recover-QO-from-ABC})}

By the uniqueness of tensor decomposition (up to column permutation and scaling) the columns of $C$
are proportional to the columns of $O$ (up to some hidden state permutation), and each column of $O$
must satisfy the normalization constraint: $\mb e^\top O_{[:,i]} = 1, \fa i\in[k]$. The
normalization in \eqref{eq:find-O} recovers $O$ from $C$.

Recall that
  \begin{align*}
    A = A^{(n)} = \Big(O\odot  A^{(n-1)}\Big) Q.
  \end{align*}
  Since the matrix $A$ has full column rank $k$, the matrices $Q\in\R^{k\times k}$ and $(O\odot
  A^{(n-1)})\in\R^{d^n\times k}$ both have full column rank $k$, as well as the pseudo-inverse of
  $(O\odot \wt A)$, therefore $ Q = (O\odot A^{(n-1)})^{\dag} A$.

  By definition we have $ A^{(1)} = OQ$, thus if $O$ is of full column rank $k$, we can obtain $Q =
  O^{\dag} A^{(1)} $.

  \qed

\tb{\noindent (Proof of Theorem~\ref{thm:correct-alg2})}

  Denote the minimal order HMM realization by $\ta^h=(k,Q,O)$, and since $n=1$, the
  matrices are given by:
  \begin{align*}
    A = OQ, \quad B = O\wt Q, \quad C = ODiag(\pi).
  \end{align*}
  Define two linear operators $I_{d^2\times d^2}: \mbb R^{d^2}\to \mbb R^{d^2}$ and $P_{d^2\times
    d^2}: \mbb R^{d^2}\to \mbb R^{d^2}$, such that for any matrix $X\in\mbb R^{d\times d}$: $
  I_{d^2\times d^2} \tx{vec}(X) = \tx{vec}(X)$ and $ P_{d^2\times d^2} \tx{vec}(X) = \tx{vec}(X^\top
  )$.
  Moreover, define matrix $R\in\mbb R^{d^2\times d^2}$ and $Q \in\mbb R^{d^4\times d^2}$ to be:
    \begin{align*}
      &R = I_{d^2\times d^2} - P_{d^2\times d^2}, \quad G = R\odot R.
    \end{align*}
    Note that the kernel of $(I_{d^2\times d^2} - P_{d^2\times d^2})$ is the space of symmetric matrices, thus
    $R$ is of rank $d^2 - d(d+1)/2 = d(d-1)/2$, and $G$ is of rank $ d^2(d-1)^2/4$.
    Define matrix $G^{\perp}\in\R^{d^4 \times (d^4- {d^2(d-1)^2\over 4})}$ such that its columns are
    orthogonal to the columns of $G$.

    According to \cite{de2006link,de2007fourth,jiang2004kruskal}, there are two deterministic
    conditions for Algorithm~\ref{alg:2} to correctly recover the factors $A,B,C$ from the rank $k$
    tensor $M$:
  \begin{enumerate}[leftmargin=*]
  \item Both $A\odot B$ and $C$ have full column rank $k$.
  \item Define $T\in \mbb R^{d^4\times (m+ (k-1)k/2)}$ to be:
    \begin{align*}
      & T =\Big[G^{\perp}_{[:,i]}: 1\le i\le d^4 - {d^2(d-1)^2\over 4},
      \\
      &A_{[:,k_1]}\odot A_{[:,k_2]}\odot B_{[:,k_1]}\odot B_{[:,k_2]}: 1\le k_1< k_2\le k \Big].
    \end{align*}
    The columns of $T$ are linear independent.
  \end{enumerate}

  Parameterize the rank $r$ transition matrix by $Q = UV^\top $ for some matrices $U,V\in\mbb
  R^{k\times r}$. Define the parameter space $\mc Q$:
  \begin{align*}
    \mc Q = \{ Q\in \mbb R^{k\times k}: Q=UV^\top , U,V\in\mbb R^{k\times r}, \mb e^\top  Q= \mb e^\top \}
  \end{align*}

  Note that by construction, the minors of $A\odot B$ and $T$ are nonzero polynomials in the
  elements of the parameters $U,V$ and $O$, in order to show that the two deterministic rank
  conditions are satisfied for almost all instances in the class $\Ta_{(d,k,r)}^h$, it is enough to
  construct an instance in the model class that satisfies the two conditions (by the random check in
  Algorithm~\ref{alg:random-check}). Moreover, if it is true, then with probability one, the two
  conditions are satisfied for a randomly chosen instance in the model class.

\qed

\section{Auxiliary lemmas}
\label{sec:auxiliary-lemmas}

\tb{(Matrix perturbation bounds)}

Since the algorithms we have examined are all based on different forms of matrix decomposition.  Characterizing the
sample complexity boils down to analyzing the stability of the matrix decompositions. Here we review some well-known
matrix perturbation bounds and prove some corollaries.

Given a matrix $\wh A = A+ E$ where $E$ is a small perturbation, the following results bound the deviation of the
singular vectors and singular values.

\begin{lemma}[Mirsky's theorem]
  \label{lem:mirsky}
  Given matrices $A,E\in\R^{m\times n}$, with $m\ge n$, then
  \begin{align*}
    \sqrt{ \sum_{i=1}^{n} (\sigma_i(A+E) - \sigma_i(A) )^2 }
    \le{ \|E\|_F }.
  \end{align*}
\end{lemma}

\begin{lemma}
  \label{lem:wedin}
  Given matrices $A,E\in\R^{m\times n}$, with $m\ge n$. Suppose that the matrix $A$ has full column
  rank and $\sigma_k(A)>0$. Let $A = USV^\top$ be the singular value decomposition of $A$, and let
  $\wh U$ and $\wh V$ denote the first $k$ left and right singular vectors of $\wh A$, let $\wh S$
  be the diagonal matrix with the first $k$ singular values of $\wh A$.
  We have:
    \begin{align*}
      &\|\wh U - \wt U\|\le {\sqrt{2} \|E\|_F\over \sigma_k(A)},
      \quad
       \|\wh V -  V\|\le {\sqrt{2} \|E\|_F\over \sigma_k(A)}.
    \end{align*}
\end{lemma}
This is an immediate corollary of Wedin's theorem.

\begin{lemma}
  \label{lem:prod-perturb}
  Consider a product of matrices $A_1\cdots A_k$, and consider any sub-multiplicative norm on matrix $\|\cdot\|$.
  Given $\wh A_1,\dots, \wh A_k$ and assume that $\|\wh A_i-A_i\|\le \|A_i\|$, then we have:
  \begin{align*}
    \|\wh A_1\cdots \wh A_k - A_1\cdots A_k\| \le 2^{k-1} \prod_{i=1}^k \|A_i\|  \sum_{i=1}^{k} {\|\wh A_i-A_i\|  \over \|A_i\|}.
  \end{align*}

\end{lemma}

\tb{(Concentration bounds)}

\begin{lemma}[Hoeffding's inequality]
  \label{lem:hoeffding}
 Let $X_1, \dots, X_n$  be independent random variables. Assume that $X_i$'s are bounded almost surely, namely
 $\Pr[X_i\in[a_i,b_i]]=1$. Define the empirical mean of these variables $\ol X = (X_1+\cdots +X_n)/n$. We have
 \begin{align*}
   \Pr[|\ol X- \mbb E[\ol X]|\ge t]\le \ep(-{2n^2 t\over \sum_{i=1}^{n} (b_i-a_i)^2 }).
 \end{align*}
\end{lemma}

\begin{lemma}[Multiplicative Chernoff bound]
  \label{lem:chernoff}
 Suppose $X_1,\cdots, X_n$ are independent random variables with Bernoulli distribution, and $\mbb
 P(X_i=1) = \mu$. Then for any $\delta>1$:
 \begin{align*}
   \mbb P\left(\sum_{i=1}^n X_i >\delta n\mu\right) < \left({e\over \delta}\right)^{\delta n\mu}.
 \end{align*}
\end{lemma}

\begin{lemma}[High dimensional sphere projection (Johnson Lindenstrauss lemma)]
  \label{lem:sphere-proj}
  Let the random vector $\mb u\in\mbb R^{d}$ be uniformly  distributed on the surface of the
  $d$-dimensional unit sphere,i.e. uniform distribution in the set:
 $  \left\{ \sum_{i=1}^{d} u_i^2 = 1\right\}$.
 Denote its projection onto the first dimension to be $|u_{1}|$. We have:
 \begin{align*}
   \mbb P(|u_{1}|>t) < {4\over \sqrt{d-2}} e^{-{d-2\over 2}t^2}.
 \end{align*}

\end{lemma}


\end{document}

%% file: test.tex
\usepackage[usenames,dvipsnames]{color}
\usepackage{stmaryrd} \usepackage{url} \usepackage[latin1]{inputenc}
\usepackage{graphicx}
\usepackage{amssymb} \usepackage{subfigure} \usepackage{amsmath}
\usepackage{amsthm}
\usepackage{dcolumn}
\usepackage{bm}
\usepackage{color}

\usepackage{algorithm} \usepackage{algorithmic}

\newtheorem{theorem}{Theorem}
 
\newtheorem{lemma}{Lemma}
\newtheorem{remark}{Remark}

\newtheorem{definition}{Definition}

\newcommand{\mc}{\mathcal}
\newcommand{\mb}{\mathbf} \newcommand{\mbb}{\mathbb} \newcommand{\wt}{\widetilde}
\newcommand{\fa}{\forall}  \newcommand{\tb}{\textbf}
\newcommand{\tx}{\text}

\newcommand{\ot}{\otimes}

\newcommand{\ta}{\theta}
\newcommand{\Ta}{\Theta}

\newcommand{\R}{\mbb R}

\newcommand{\ol}{\overline}
\newcommand{\wh}{\widehat}

\newcommand{\ep}{\tx{exp}}
\newcommand{\lt}{\left}
\newcommand{\rt}{\right}